\pdfoutput=1

\documentclass[11pt]{article}

\usepackage{EMNLP2023}

\usepackage{times}
\usepackage{latexsym}

\usepackage[T1]{fontenc}

\usepackage[utf8]{inputenc}

\usepackage{microtype}

\usepackage{inconsolata}

\usepackage{lipsum}
\usepackage{tabularx}
\usepackage{mdframed}
\definecolor{agentcolor}{HTML}{165FE0}
\definecolor{clientcolor}{HTML}{00830D}

\usepackage{booktabs} 
\usepackage{todonotes} 

\usepackage{amsmath}     
\usepackage{amssymb}     
\usepackage{mathtools}   
\usepackage{dsfont}      
\usepackage{xcolor, colortbl}
\usepackage{listings}
\title{Synthetic Dialogue Dataset Generation using LLM Agents}

\author{Yelaman Abdullin \and Diego Molla-Aliod \\
        Macquarie University \\
        \texttt{yelaman.abdullin@hdr.mq.edu.au}, \texttt{diego.molla-aliod@mq.edu.au}
        \AND
        Bahadorreza Ofoghi \and John Yearwood \\
        Deakin University\\
        \texttt{\{b.ofoghi,john.yearwood\}@deakin.edu.au}
        \And
        Qingyang Li\\
        The University of Melbourne\\
        \texttt{ql5@student.unimelb.edu.au}
        }

\begin{document}
\maketitle
\begin{abstract}
Linear programming (LP) problems are pervasive in real-life applications. However, despite their apparent simplicity, an untrained user may find it difficult to determine the linear model of their specific problem. We envisage the creation of a goal-oriented conversational agent that will engage in conversation with the user to elicit all information required so that a subsequent agent can generate the linear model. In this paper, we present an approach for the generation of sample dialogues that can be used to develop and train such a conversational agent. Using prompt engineering, we develop two agents that ``talk'' to each other, one acting as the conversational agent, and the other acting as the user. Using a set of text descriptions of linear problems from NL4Opt available to the user only, the agent and the user engage in conversation until the agent has retrieved all key information from the original problem description. We also propose an extrinsic evaluation of the dialogues by assessing how well the summaries generated by the dialogues match the original problem descriptions. We conduct human and automatic evaluations, including an evaluation approach that uses GPT-4 to mimic the human evaluation metrics. The evaluation results show an overall good quality of the dialogues, though research is still needed to improve the quality of the GPT-4 evaluation metrics. The resulting dialogues, including the human annotations of a subset, are available to the research community. The conversational agent used for the generation of the dialogues can be used as a baseline.
\end{abstract}

\section{Introduction}
Linear programming (LP) is a mathematical optimization technique widely employed to address a multitude of real-world challenges, ranging from resource allocation in supply chain management to portfolio optimization in finance. Despite the ubiquity of LP problems and their practical importance, many individuals, particularly those without specialized mathematical backgrounds, often struggle to formulate the appropriate linear models for their specific problem instances. This barrier hinders the broader utilization of LP techniques, especially among non-experts.

To mitigate this challenge, we propose the development of a goal-oriented conversational agent capable of assisting users in constructing accurate linear models for their unique problem scenarios. This conversational agent would engage users in a dialogue, eliciting relevant information pertaining to the problem, and subsequently generate the corresponding linear model. This paper focuses on an essential aspect of creating such an agent –-- the generation of synthetic dialogues that can be employed to train and evaluate the conversational agent's performance.

Our methodology leverages prompt engineering to construct two distinct agents: one simulating the conversational agent's behavior, and the other emulating the user's responses during problem-solving interactions. The agents are designed to engage in purposeful dialogues aimed at extracting the necessary information from the user to construct a valid linear model. To facilitate this process, we utilize a set of text descriptions of linear problems, accessible only to the user agent, sourced from the NL4Opt dataset \cite{ramamonjison-etal-2022-augmenting, ramamonjison2023nl4opt}. These text descriptions serve as the basis for the dialogues and enable the conversational agent to iteratively gather the critical information required for problem formulation.

In addition to the generation process, we propose an extrinsic evaluation approach for assessing the quality and effectiveness of the generated dialogues. Specifically, we evaluate how well the summaries generated by the dialogues align with the original problem descriptions from NL4Opt. This evaluation encompasses both human assessments, where human judges evaluate the quality of dialogue summaries, and automated metrics to quantitatively measure the informativeness of the generated summaries.

Our preliminary results from human and automatic evaluations indicate that the generated dialogues exhibit a high degree of fidelity to the original problem descriptions, thereby demonstrating the quality of the synthetic dialogues generated. 

The contributions of this paper are:
\begin{enumerate}
    \item An approach for the generation of dialogues for the development of goal-oriented conversational agents. In this paper, the goal consists of eliciting information from the user in order to generate a linear programming model, noting that the techniques presented here can be adapted to other goals.
    \item A dataset of 476 dialogues for the development of such a conversational agent, of which 28 have been annotated manually.\footnote{\url{https://github.com/eabdullin/optimouse-quest/}} Even though these 476 dialogues are generated automatically, since the generation process is non-deterministic, separate runs of the same program will generate different dialogues. For this reason, we consider that these dialogues form a useful dataset for the research community to facilitate reproducibility. 
    \item An extrinsic evaluation approach based on comparing the summaries generated by the dialogue, with the original problem description.
    \item An automatic evaluation approach using GPT-4 that mimics the behavior of human evaluation.
\end{enumerate}


\section{Background and Related Work}

\subsection{Linear Programming and NL4Opt}

LP problems are pervasive in real-life applications. They are commonly utilized for resource allocation, planning, scheduling, transportation optimization, portfolio management, and numerous other areas. For instance, in production planning, LP can help determine how to use limited human, material, and financial resources to achieve maximum economic benefits. 

LP problems are a class of mathematical optimization problems where the goal is to find a set of values for the decision variables that satisfies a set of linear constraints and maximizes or minimizes the value of a linear objective function \cite{chen11applied}. The general form of an LP problem can be formulated as follows, 
\begin{equation*}
\begin{array}{ll@{}ll}
\text{Maximize}  & \mathbf{c}^T &\mathbf{x} &\\
\text{subject to}& \mathbf{A}   &\mathbf{x} \leq \mathbf{b}\\
                 &              &\mathbf{x} \geq \mathbf{0}
\end{array}
\end{equation*}
where $\mathbf{x}$ is a vector of decision variables $\in \mathbb{R}^n$, $\mathbf{c}$ and $\mathbf{b}$ are given vectors of constants, and $\mathbf{A}$ is a given matrix of constants.
Linear programming is widely applicable in various domains due to its ability to model real-world optimization challenges and the availability of efficient solution algorithms. Once an LP model has been formulated for a problem, powerful solvers driven by efficient algorithms can help one to solve it, even for surprisingly complicated and large-scale problems.

In 2022, a competition to extract linear programming formulations from natural language (NL4Opt) developed the first dataset of linear programming word problems \cite{ ramamonjison-etal-2022-augmenting, ramamonjison2023nl4opt}. It contains 1101 instances from various domains. These LP problems focus on a variety of common contexts such as production, resource allocation, investment allocation, agriculture, transportation, health sciences, sales, etc. Each instance has an unstructured natural language description of the LP problem involving decision variables, one objective function, and several constraints.

The NL4Opt dataset provides valuable examples of real-world natural language descriptions for LP problems, showcasing a range of constraint types. We deconstruct each problem description in the development subset into an objective function description and several constraint descriptions. By analyzing these constraints, we find that they can be categorized into nine refined classes in Table~\ref{tab:constraint class}. Different instances of the dataset consist of different combinations of these constraint types. Constraint types 1 to 3 and 5 to 7 are frequently used to represent capacity limits, budget constraints, or resource availability. Types 4, 8, and 9 impose ratio control and balancing between different quantities. These nine types of constraints are often encountered in real-world problems and can encompass a substantial portion of common constraints.

\begin{table}[!h]
    \small
    \begin{tabular}{lp{3.9cm}l}
    &Constraint type & Math inequality\\
    \midrule
    1 & Upper bound on single variable & $ x_i \leq b $\\
    2 & Upper bound on sum of variables & $\sum_{i} x_i \leq b $\\
    3 & Upper bound on weighted sum of variables & $\sum_{i} a_i x_i \leq b $\\
    4 & Upper bound on proportion & $ x_j \leq c \sum_{i} x_i $\\ 
    5 & Lower bound on single variable & $ x_i \geq b $\\
    6 & Lower bound on sum of variables & $\sum_{i} x_i \geq b $ \\ 
    7 & Lower bound on weighted sum of variables & $\sum_{i} a_i x_i \geq b $\\
    8 & Lower bound on proportion & $ x_j \geq c \sum_{i} x_i $\\
    9 & Comparison constraints & $ d x_i \leq x_j $\\
    \end{tabular}
    \caption{Classification of constraints. Suppose an LP problem has $n$ decision variables, $x_i$ and $x_j$ are decision variables, $a_i$ and $b$ are nonnegative constants, $d$ is a positive constant, and $c$ is a constant $\in (0, 1]$, $i, j \in I=\{1,2, \ldots, n\}$.}\label{tab:constraint class}
\end{table}

\subsection{Evaluation Methods}
Evaluation of dialogue systems is a complex endeavor, typically involving a blend of automated metrics and human assessments. A traditional automated metric such as ROUGE \cite{lin-2004-rouge} is frequently employed for measuring textual similarity and evaluating information overlap. However, this metric could be improved in its ability to evaluate semantic coherence and the effective fulfillment of dialogue goals \cite{liu-etal-2016-evaluate}. In this work, we calculated ROUGE-1, ROUGE-2, and ROUGE-L scores to cover unigram, bigram, and longest common subsequence overlaps, respectively.

Recent advancements have directed more sophisticated evaluation methods, including BERTScore \cite{zhang2020bertscore}, which leverages contextual embeddings from pre-trained BERT models to assess semantic similarity between generated and reference text. BERTScore complements ROUGE by adding a more nuanced semantic layer to the evaluation, capturing aspects that traditional metrics may miss.

Moreover, the rise of Large Language Models (LLMs) has further enriched the toolkit for text generation evaluation  \cite{fu2023gptscore, liu2023geval}. LLMs, with their ability for nuanced analysis, offer possibilities beyond syntactic and surface-level metrics. They can potentially capture deeper aspects of dialogue semantics and goal alignment.

\section{Methodology}
This section outlines the methods employed to accomplish two objectives of this study: automating the generation of dialogue datasets, and evaluating the quality of the generated dialogues. We utilize a dual-agent setup leveraging LLM, in our case, OpenAI's GPT-4 \cite{OpenAI2023GPT4TR}, to simulate a conversation between a user and an assistant focusing on linear programming problems.

\subsection{Dual-Agent LLM Setup}

The dual-agent setup aims to model a conversation between an automatic conversational agent who asks questions to a user, with the aim to identify all key information of the linear problem to model. Specifically, the setup comprises two distinct LLM agents: a Question Generation Agent and a Question Answering Agent who ``talk'' to each other.

\begin{itemize}
    \setlength\itemsep{0em}
    \item \textbf{Question Generation (QG) Agent}: The primary objective of this agent is to ask questions that will enable it to elicit and summarize the key information of the original problem statement. Importantly, this agent does not have access to the original problem statement and must rely solely on its interactions with the QA Agent to gather all necessary information.
    \item \textbf{Question Answering (QA) Agent}: This agent is designed to answer questions based on a pre-defined problem statement from NL4Opt, which serves as the simulated knowledge base for the assistant. To make the dialogue more natural and contextually grounded, the agent is configured to impersonate the individual mentioned in the original problem statement. This design choice enhances the genuineness of the interaction, creating a more realistic user-agent dialogue experience.
\end{itemize}

\subsection{Implementation Details}

An essential part of the QA Agent is a component (which also leverages LLM) that compares a summary with an original problem statement. This component's role is twofold: to provide feedback when discrepancies are found between the provided summary and the original problem statement, and to signal the termination of dialogue generation if not. To detect that the QG has generated a summary in the latest dialogue turn, we employed a straightforward approach using regular expressions and predefined keywords.

In order to maintain consistency and guide the LLM in its responses, prompts were injected throughout the dialogue. Aside from the initial system prompt, which always begins the conversation, subsequent prompts were introduced at each turn and retained in the dialogue's history. For the QG Agent, a succinct instruction, \textit{A NEXT MESSAGE/QUESTION} was provided. For the QA Agent, the prompts varied based on the dialogue stage. Mostly, the instruction \textit{ANSWER SHORTLY. USE MAXIMUM 30 WORDS.} was used. However, when the QG agent provided a summary, the prompt adjusted accordingly: either feedback on summary\footnote{Full prompts are provided in Appendix~\ref{sec:appendix_llm_prompts}.}, or \textit{THE SUMMARY ACCEPTED. IT'S TIME TO FINISH DIALOG AND SAY GOODBYE}, signaling the end of the conversation. 

We utilized the dataset of linear programming problems from the NL4Opt competition \cite{ramamonjison2023nl4opt}, specifically from subtask 2. The task is designed to take a natural language problem description, labeled semantic entities, and an ordered mapping of variable mentions, and then formulate a precise meaning representation. These problem descriptions served as the basis for the simulated conversations between the two LLM agents in this work.

\begin{itemize}
    \setlength\itemsep{0em}
    \item \textbf{Development Subset}: For this subset, consisting of 98 problems. Two to four separate dialogues were generated per problem. 
    \item \textbf{Training Subset}: For the training subset, a single dialogue was generated for the first 241 problems.
\end{itemize}

The dual-agent system was implemented using OpenAI's Chat Completion API\footnote{\url{https://platform.openai.com/docs/api-reference/chat}}. We generated the dialogues using different temperature settings for the development subset to ensure a variety of dialogues. The aim was to generate multiple dialogue instances for each problem so that it is possible to robustly evaluate the model's performance and assess the diversity in the generated dataset in subsequent research.

\begin{figure}[h!]
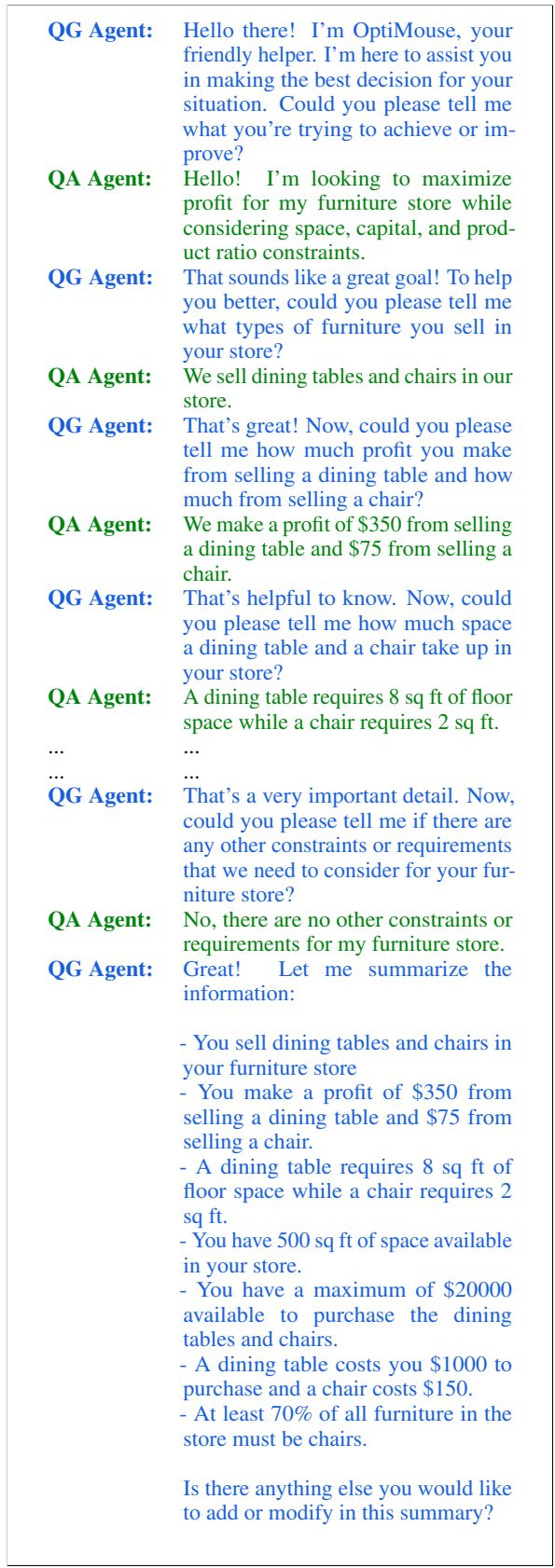

    \centering
    \begin{mdframed}
        \small
        \begin{tabularx}{\linewidth}{l X}
            \textcolor{agentcolor}{\textbf{QG Agent:}} & \textcolor{agentcolor}{Hello there! I'm OptiMouse, your friendly helper. I'm here to assist you in making the best decision for your situation. Could you please tell me what you're trying to achieve or improve?} \\
            \textcolor{clientcolor}{\textbf{QA Agent:}} & \textcolor{clientcolor}{Hello! I'm looking to maximize profit for my furniture store while considering space, capital, and product ratio constraints.} \\
            \textcolor{agentcolor}{\textbf{QG Agent:}} & \textcolor{agentcolor}{That sounds like a great goal! To help you better, could you please tell me what types of furniture you sell in your store?} \\
            \textcolor{clientcolor}{\textbf{QA Agent:}} & \textcolor{clientcolor}{We sell dining tables and chairs in our store.} \\
            \textcolor{agentcolor}{\textbf{QG Agent:}} & \textcolor{agentcolor}{That's great! Now, could you please tell me how much profit you make from selling a dining table and how much from selling a chair?} \\
            \textcolor{clientcolor}{\textbf{QA Agent:}} & \textcolor{clientcolor}{We make a profit of \$350 from selling a dining table and \$75 from selling a chair.} \\
            \textcolor{agentcolor}{\textbf{QG Agent:}} & \textcolor{agentcolor}{That's helpful to know. Now, could you please tell me how much space a dining table and a chair take up in your store?} \\
            \textcolor{clientcolor}{\textbf{QA Agent:}} & \textcolor{clientcolor}{A dining table requires 8 sq ft of floor space while a chair requires 2 sq ft.} \\
            ... & ...\\
            ... & ...\\
            \textcolor{agentcolor}{\textbf{QG Agent:}} & \textcolor{agentcolor}{That's a very important detail. Now, could you please tell me if there are any other constraints or requirements that we need to consider for your furniture store?} \\
            \textcolor{clientcolor}{\textbf{QA Agent:}} & \textcolor{clientcolor}{No, there are no other constraints or requirements for my furniture store.} \\
            \textcolor{agentcolor}{\textbf{QG Agent:}} & \textcolor{agentcolor}{Great! Let me summarize the information:\newline\newline\quad- You sell dining tables and chairs in your furniture store\newline\quad- You make a profit of \$350 from selling a dining table and \$75 from selling a chair.\newline\quad- A dining table requires 8 sq ft of floor space while a chair requires 2 sq ft.\newline\quad- You have 500 sq ft of space available in your store.\newline\quad- You have a maximum of \$20000 available to purchase the dining tables and chairs.\newline\quad- A dining table costs you \$1000 to purchase and a chair costs \$150.\newline\quad- At least 70\% of all furniture in the store must be chairs.\newline\newline Is there anything else you would like to add or modify in this summary?}

        \end{tabularx}
    \end{mdframed}
    \caption{A sample dialogue between a QG Agent and a QA Agent about maximizing furniture store profits.}
    \label{fig:conversation}
\end{figure}

Figure~\ref{fig:conversation} shows part of a sample dialogue, and figure ~\ref{fig:problemstatement} shows the associated problem statement. Appendix~\ref{sec:appendix_llm_prompts} shows the actual prompts used.

\begin{figure}[h!]
    \centering
    
    \begin{mdframed}
        \small
        \textit{A furniture store only stocks and sells dining tables and chairs. The profit per dining table is \$350 and the profit per chair is \$75. There is 500 sq ft of space available and a dining table requires 8 sq ft of floor space while a chair requires 2 sq ft. Because chairs sell in larger quantities, at least 70\% of all furniture in the store must be chairs. In terms of capital, a dining table ties up \$1000 in capital and a chair ties up \$150 in capital. The company wants a maximum of \$20000 worth of capital tied up at any time. Formulate an LP to maximize profit.}
    \end{mdframed}
    \caption{A sample problem statement from the NL4Opt dataset.}
    \label{fig:problemstatement}
\end{figure}

\section{Evaluation Strategy} 

To evaluate the quality of the generated dialogues, we conduct an \emph{extrinsic evaluation}.
The evaluation process focuses on an underlying hypothesis that serves as the core of our study: \textit{a correct summary generated by the QG Agent would imply that the agent has posed the appropriate and necessary questions to restore the original problem statement effectively}. The hypothesis is grounded in several key assumptions and logical deductions:

\vspace{6pt}\noindent  \textbf{Extrinsic evaluation}. In dialogue systems, particularly goal-oriented systems, the dialogue often progresses through a series of questions and answers to reach a particular end state \cite{qi-etal-2020-stay, majumder-etal-2021-ask}. Therefore, the questions asked directly impact the quality and accuracy of the final output --- here, the summary of the original problem statement. We, therefore, conduct an extrinsic evaluation of the dialogues.

\vspace{6pt}\noindent \textbf{Incomplete Initial Information}. The QG Agent starts with incomplete information about the problem at hand. Therefore, asking the right questions is crucial for the agent to gather enough information for an accurate summary. A poor or incomplete summary would suggest that the agent has asked insufficient or incorrect questions. 

\vspace{6pt}\noindent\textbf{Semantic Understanding}. An accurate summary involves not just factual correctness but also a semantic understanding of the problem's constraints and objectives. Therefore, correct summarization can be viewed as an implicit validation of the agent's ability to grasp the problem's complexities through its questions.

\vspace{6pt}\noindent To validate these assumptions, we use an approach involving human evaluation as well as automated evaluation through a third LLM agent of generated summaries. By evaluating the correctness of the generated summaries, we thus indirectly assess the QG Agent's ability to pose appropriate and informative questions that lead to a comprehensive understanding of the problem.


\subsection{Automatic Evaluation}\label{sec:automatic_evaluation}

For the automated evaluation, in addition to the well-known ROUGE and BERTScore, we designed a metric that employed a third LLM (GPT-4) agent to compare the original problem statement with the generated summary. The evaluation was conducted using a ``chain-of-thought'' prompting \cite{NEURIPS2022_9d560961}. Our agent employs three criteria (``chains'') to evaluate each summary:

\begin{itemize}
    \setlength\itemsep{0em}
    \item \textbf{Correct Information}: Assessing if the summary accurately represents the facts in the original problem statement.
    \item \textbf{Incorrect Information}: Identifying any factual inaccuracies or misleading statements in the summary.
    \item \textbf{Missing Information}: Determining if any crucial elements from the original problem statement are bypassed in the summary.
\end{itemize}

Considering the three criteria above, the agent provides an "Information Recall Score", "Information Precision Score", "Information Repetition Score", and "Readability Score" to judge if the summary forms a coherent and accurate representation of the original problem. These evaluation metrics are the same as the human evaluation conducted in Section~\ref{sec:human_evaluation}. Appendix~\ref{sec:appendix_llm_prompts} shows the prompts used.

\subsubsection{Results of the Automatic Evaluation}

\begin{table}[]
    \centering
    \begin{tabular}{lr}
    Metric & Value\\
    \midrule
ROUGE-1 P & 0.54\\
ROUGE-1 R & 0.62\\
ROUGE-1 F1 & 0.57\\
\midrule
ROUGE-2 P & 0.33\\
ROUGE-2 R & 0.39\\
ROUGE-2 F1 & 0.35\\
\midrule
ROUGE-L P & 0.38\\
ROUGE-L R & 0.43\\
ROUGE-L F1 & 0.40\\
\midrule
BERTscore P & 0.88\\
BERTscore R & 0.91\\
BERTscore F1 & 0.90\\
\midrule
GPT-4 R & 4.60\\
GPT-4 P & 4.62\\
\end{tabular}
    \caption{Average values of metrics per entire dataset. All values range from 0 to 1 except GPT-4, which ranges from 1 to 5.}
    \label{tab:automatic_eval_aggregate}
\end{table}

Table \ref{tab:automatic_eval_aggregate} summarizes the average metric values across the entire dataset. As we can observe, the absolute values of ROUGE are not very high. However, as we will see below, human evaluation of a subset of the data reveals that the dialogues have generated good summaries in general.

\begin{table}[]
    \centering
    \begin{tabular}{lr}
    Name & Value\\
    \midrule
Total number of dialogues & 476 \\
Dialogues with temperature 0 & 315\\
Dialogues with temperature 1 & 149\\
Dialogues with summary & 97\%\\
Average dialogue length (characters) & 3658\\
Total number of turns & 9480\\
Average number of turns per dialogue & 20 \\
Average turn length (characters)& 184\\
    \end{tabular}
    \caption{Summary statistics of generated dialogues.}
    \label{tab:dialogue_summary_stats}

\end{table}

To provide a more nuanced understanding of our generated dialogues, we have collected some summary statistics. These are presented in Table \ref{tab:dialogue_summary_stats}. The dialogues were generated with varying temperature settings to control the randomness of the text generated by the LLMs. In a small number of cases (3\%), the dialogue was not able to generate a summary before the cut-off of 40 turns (20 turns for each agent). The high percentage of dialogues with a summary (97\%) indicates the Question Generation Agent's effectiveness in concluding the dialogues with a summary, which is crucial for our evaluation. Dialogue length and number of turns give an indication of the depth and extent of the conversations between the agents.

\subsection{Human Evaluation}\label{sec:human_evaluation}

For the human evaluation component, we carefully curated a subset of 28 problem statements from the development subset. This subset was selected to cover all constraint types represented in the development data (Table~\ref{tab:constraint class}), ensuring a comprehensive evaluation across diverse problem scenarios.

Given the small base of problems (98), the selection of 28 problems was simple: first, apply a greedy approach that satisfies the quantity requirements of the types with fewer counts, and then manually swap selected candidates with other candidates from the development set as appropriate, until we reached an acceptable distribution of constraint types in the selection. Table~\ref{tab:selectiondistributions} shows the distribution of constraint types in the development set and the selection used for human evaluation.

\begin{table}[]
    \centering\small
    \begin{tabular}{lp{0.6\linewidth}rr}
&Constraint Type        &  Dev & Sel\\
\midrule
1&Upper bound on single variable & 20 & 6\\
2&Upper bound on sum of variables & 12 & 4\\
3&Upper bound on weighted sum of variables & 93 & 28 \\ 
4&Upper bound on proportion & 8 & 2\\
5&Lower bound on single variable & 35 & 11\\
6&Lower bound on sum of variables & 7 & 2\\
7&Lower bound on weighted sum of variables & 59 & 18 \\
8&Lower bound on proportion & 15 & 5\\
9&Comparison & 43 & 13\\
    \end{tabular}
    \caption{Counts of types of linear optimization constraints in the development set of 98 problems (``Dev'' column), and the selection of 28 problems (``Sel'' column). The sum is larger than the number of problems since a problem may have multiple constraints.}
    \label{tab:selectiondistributions}
\end{table}

To add an element of variability in dialogue generation, each problem statement in this selection was subjected to a single dialogue generation run. The temperature setting for this run was randomly selected to be either 0 or 1.


Four evaluators (details in the Acknowledgements section) then scored how well the summary generated at the end of the dialogue matches the problem statement. For every pair of a problem statement and a generated summary, each evaluator produced the following 4 evaluation metrics. These metrics have been adopted from the human evaluation performed by \citet{Tsatsaronis:2015}:

\begin{itemize}
\item \textbf{Information recall (IR)} (1-5) -- All the necessary information is in the generated summary.
\item \textbf{Information precision (IP)} (1-5) -- No irrelevant information is generated.
\item \textbf{Information repetition (IRep)} (1-5) -- The generated summary does not repeat the same information multiple times.
\item \textbf{Readability (Read)} (1-5) -- The generated summary is easily readable and fluent.
\end{itemize}

\subsubsection{Results of the Human Evaluation}

Inter-annotator agreement of each of the 4 human evaluation metrics, as computed by Fleiss' Kappa, is shown in Table~\ref{tab:kappa}.

\begin{table}[]
    \centering
    \begin{tabular}{lr}
    Metric & Fleiss' Kappa\\
    \midrule
         Information Recall & 0.205\\
         Information Precision & 0.387\\
         Information Repetition & -0.009\\
         Readability & 0.235\\
    \end{tabular}
    \caption{Inter-annotator agreement of each of the 4 human evaluation metrics.}
    \label{tab:kappa}
\end{table}

We observe virtually no agreement in Information Repetition, slight agreement in Information Recall and Readability, and Fair agreement in Information Precision.

Table~\ref{tab:human} shows the average values of the human evaluation scores for the selection of 28 problems.
\begin{table}[]
    \centering
    \begin{tabular}{crrrr}
        Annotator & IR & IP & IRep & Read \\
        \midrule
         1 & 4.25 & 4.25 & 4.89 & 4.96\\
         2 & 4.18 & 4.54 & 4.93 & 4.96\\
         3 & 4.68 & 4.39 & 4.93 & 4.86\\
         4 & 4.03 & 4.36 & 4.82 & 4.89\\
         \midrule
         All & 4.29 & 4.38 & 4.89 & 4.92
    \end{tabular}
    \caption{Average human evaluation scores for the sample of 28 documents. IR = Information Recall. IP = Information Precision. IRep = Information Repetition. Read = Readability.}
    \label{tab:human}
\end{table}
Overall, the human evaluation showed high values, and very high values for Information Repetition and Readability. This suggests that the dialogues generated by the pair of agents are of good quality.

The human annotators observed the following most common mistakes in the generated summaries. See Appendix~\ref{sec:appendix_mistakes} for examples of each.

\begin{itemize}
    \item Missing objective function or decision variables.
    \item Inclusion of additional information that seems to be from answers to the agent asking clarifying questions, such as ``otherwise the order does not matter'', or ``there is no upper limit on costs'', etc.
\end{itemize}

\subsection{Correlation Analysis of Automatic and Human Evaluations}

Table~\ref{tab:humancomputercorrelation} shows Spearman's rank correlation coefficient $\rho$ between the automatic and human evaluations. 
\begin{table}[]
    \centering
    \begin{tabular}{lrrrr}
      Metric &  IR & IP & IF1 & IAvg\\
      \midrule
      ROUGE-1 R & 0.43&&\\
      ROUGE-1 P &&0.58&\\
      ROUGE-1 F1 &&&0.62&0.60\\
      \midrule
      ROUGE-2 R &0.48&&\\
      ROUGE-2 P &&0.58&\\
      ROUGE-2 F1 &&& 0.56&0.57\\
      \midrule
      ROUGE-L R &0.47&&\\
      ROUGE-L P &&0.74&\\
      ROUGE-L F1 &&& 0.71&0.69\\
      \midrule
      BERTScore R &0.53&&\\
      BERTScore P &&0.74&\\
      BERTScore F1 &&& 0.65&0.65\\
      \midrule
      GPT-4 R & 0.42&&\\
      GPT-4 P && 0.67&\\
      GPT-4 F1 &&& 0.59 & 0.58\\
    \end{tabular}

    \caption{Spearman's rank correlation coefficient $\rho$ between the automatic and human evaluations. IP = correlation with Information Recall; IR = correlation with Information Precision; IF1 = Correlation with the Harmonic mean of IR and IP; IAvg = Correlation with the average of Information Recall, Information Precision, Repetition, and Readability.}
    \label{tab:humancomputercorrelation}
\end{table}
The table compares recall values of the automatic metrics against Information Recall, precision values against Information Precision, and F1 values against the harmonic mean of Information Recall and Information Precision. The last column of the table shows the correlation between the F1 values of the automatic metrics and the average of Information Recall, Information Precision, Information Repetition, and Readability. We can observe nearly identical values to the harmonic mean of Information Recall and Information Precision. The reason for this may be that the annotations for Information Repetition and Readability are nearly always 5, so their contribution is almost a constant value that does not change the rankings, so they do not affect the values of $\rho$. This may be a consequence of using GPT-4 and the carefully designed prompts, which instruct the system to be clear and concise.

Among all automatic metrics, Table~\ref{tab:humancomputercorrelation} shows that the best correlation values are for ROUGE-L. GPT-4 achieved competitive results but did not outperform the other metrics. Further work is needed to improve the use of GPT-4. In particular, GPT-4 usually was more generous and would give higher ratings than the human evaluators would. The inclusion of few-shot samples, and more sophisticated prompts, might help the system align with the human annotators.

\section{Summary and Conclusion}

This paper presents a dataset for the task of eliciting information from the user through a dialogue with a conversation agent. The specific use of the information elicited is for automatic modeling of linear optimization problems. This is \emph{per se} a very useful task with broad potential applications, but the methods for data generation and evaluation proposed here can be adopted easily for other possible tasks. The data and human evaluations are available to the research community.\footnotemark[1]

The dialogue was generated in a dual-agent LLM setup where a question generation agent acted as the machine agent who elicited information, and a question answering agent acted as the human who had the information about the problem to model. The question generation agent can be used as a baseline agent. The human evaluation results indicate that this baseline may be effective for the task.

The dialogues for a subset of 28 LP problems were evaluated using an extrinsic evaluation that judged whether summaries generated by the dialogues matched the key information from the original problem descriptions. The evaluation was conducted by human evaluators and automatically. Among the automatic evaluation metrics, besides well-known automatic metrics ROUGE and BERTScore, we designed another GPT-4 agent that mimicked the human evaluators. 
The results indicate a reasonable correlation between ROUGE L, BERTScore P, and the average human information precision scores, and this is slightly better than the correlation between the  GPT4 agent and the human IP scores.

As further work, we intend to refine the prompts used for the evaluation approach with GPT-4. In addition, we will conduct more exhaustive types of evaluation on the data set that might be more suitable to the specific domain of linear programming modeling. In particular, we plan to analyse the generated dialogues at the level of the dialogue turns.

\section*{Acknowledgements}
This work was partially funded by the Australian Research Council, Australia through the Discovery Project 2022 (grant number DP220101925). In addition to two authors of this paper who have annotated the data (Diego Molla-Aliod and John Yearwood), we acknowledge the annotations provided by Vicky Mak-Hau and Thuseethan Selvarajah.

\bibliography{anthology,custom}
\bibliographystyle{acl_natbib}

\appendix

\section{Sample Generated Summaries with Mistakes}
\label{sec:appendix_mistakes}

The following generated summary did not include the objective function:

\begin{quote}
\em
\textbf{Problem statement:}\\
A suit store makes 2-piece and 3-piece suits. To make a 2-piece suit requires 1 hour of measuring and 4 hours of stitching. To make a 3-piece suit requires 2 hours of measuring and 5 hours of stitching. The store has 150 hours of measuring time and 300 hours of stitching time available. In order to stay in business, at least 40 2-piece suites should be made. The profit per 2-piece suit is \$600 and the profit per 3-piece suit is \$800. Formulate a LP to maximize profit?

\textbf{Generated summary:}\\
Alright, thanks for the information! To make sure I've got everything, here's a summary of your situation:

- You manufacture two types of suits: 2-piece suits and 3-piece suits.\\
- A 2-piece suit makes a profit of \$600, requires 1 hour of measuring time and 4 hours of stitching time.\\
- A 3-piece suit makes a profit of \$800, requires 2 hours of measuring time and 5 hours of stitching time.\\
- You have 150 hours of measuring time and 300 hours of stitching time available.\\
- At least 40 2-piece suits must be produced, but there is no minimum requirement for 3-piece suits.\\
- There is no limit on how many suits you can sell.\\

Have I missed out on anything or is there any other specific requirement or constraint that I should know about?
\end{quote}

The following generated summary did not mention decision variables:

\begin{quote}
\em
\textbf{Problem statement:}\\
A chemical company makes an acidic liquid and a basic liquid using two types of techniques, reaction X and reaction Y. Reaction X can make 35 au of the acidic liquid and 75 au of the basic liquid per minute. Reaction Y can make 42 au of the acidic liquid and 35 au of the basic liquid per minute. Furthermore, reaction X requires 18 units of heat every minute while reaction Y requires 12 units of heat every minute. The chemical company has the capacity to generate at most 1400 units of heat every minute. If the chemical company needs to make at least 900 au of the acidic liquid and 1200 au of the basic liquid per minute, how many minutes should each reaction be run for to minimize the total time needed?

\textbf{Generated summary:}\\
Great! Let's summarize the information:

- Your company has two types of reactions, X and Y, that produce acidic and basic liquids.\\
- Reaction X requires 18 units of heat per minute and produces 35 au of acidic and 75 au of basic liquid.\\
- Reaction Y requires 12 units of heat per minute and produces 42 au of acidic and 35 au of basic liquid.\\
- Your company can generate at most 1400 units of heat every minute.\\
- You need to produce at least 900 au of the acidic liquid and 1200 au of the basic liquid per minute.\\
- Your goal is to meet these production requirements in the shortest time possible.\\

Is there anything else that I should know or did I miss anything?

\end{quote}

The following summary has additional information that makes sense but is not in the original problem description:

\begin{quote}
\em
\textbf{Problem statement:}\\
A spicy paste is to be made from units of peppers and tomatoes and is to contain a minimum of 130 units of spice and 90 units of flavor. Peppers cost \$8 per unit and tomatoes cost \$5 per unit. One unit of peppers contains 12 units of spice and 2 units of flavor. One unit of tomatoes contains 1 unit of spice and 8 units of flavor. Formulate this as an LP and find the minimum cost spicy paste that can be made.

\textbf{Generated summary:}\\
Thanks for the information. Based on your responses, here's your situation as I understand:

- You're making a spicy paste using peppers and tomatoes.\\
- Each unit of peppers costs \$8, contributes 12 units of spice and 2 units of flavor.\\
- Each unit of tomatoes costs \$5, contributes 1 unit of spice and 8 units of flavor.\\
- Each batch of spicy paste should contain at least 130 units of spice and 90 units of flavor.\\
- There are no restrictions on the maximum units of peppers and tomatoes you can use.\\
- There's no specific budget, but you aim to minimize costs.

Have I captured everything correctly?
\end{quote}

\section{LLM Prompts}
\label{sec:appendix_llm_prompts}

\textbf{Evaluation Prompt:}
\begin{lstlisting}

You are an AI evaluator specializing in assessing the quality of summaries.
Carefully check how the summary captured a linear programming problem statement.
Important information for this task is explicit names and values of decision variables, constraints of all types, and an objective function.
Your primary goal is to rate the summary based on Information Recall, Information Precision, Information Repetition and Readability.

The Problem Statement:         
```
{0}
```

The Provided Summary:
'''
{1}
'''

PROVIDE THE ANSWER IN A JSON FORMAT WITH FOLLOWING FIELDS: 
"correct_information" - string | information accurately captured in the summary
"missing_information" - string | important information existing in the original problem statement but not captured in the summary.
"incorrect_information" - string | information existing in an original problem description but wrongly/incorrectly captured in a summary
"Information Recall Score" - int | Score from 1 to 5
"Information Precision Score" - int | Score from 1 to 5
"Information Repetition Score" - int | Score from 1 to 5
"Readability Score" - int | Score from 1 to 5
\end{lstlisting}
\textbf{QG Agent Prompt:}
\begin{lstlisting}
YOU ARE "OptiMouse" - A CHATBOT HELPING USERS TO FORMULATE FULL OPTIMIZATION PROBLEM STATEMENT.
THE USER IS NOT A MATH EXPERT AND HAS NO EXPERIENCE WITH MATH AND OPTIMIZATIONS.
DO NOT USE ANY MATHEMATICAL TERMINOLOGY OR EXPLANATIONS SUCH AS OBJECTIVE FUNCTION, CONSTRAINTS, ETC.

GATHER NECESSARY DETAILS THAT CAN BE MAPPED TO A LINEAR PROGRAMMING MODEL.
ENGAGE USERS BY ASKING CLEAR, CONCISE, AND SEQUENTIAL QUESTIONS TO RECEIVE INFORMATION ABOUT CONSTRAINTS AND OBJECTIVE FUNCTION.
ASK A QUESTION BASED ON THE PREVIOUS INFORMATION THAT WILL LEAD TO GETTING A CONSTRAINT OR OTHER PARAMETER OF THE MODEL.
THINK DEEPLY SO YOU WILL BE ABLE TO GET FULL PROBLEM DETAILS.
ONE QUESTION ALLOWED PER MESSAGE. 

PROVIDE A SUMMARY IN BULLET POINTS (SEE EXAMPLE DELIMITED BY "====") ONCE YOU HAVE ALL THE INFORMATION NEEDED
DO NOT INCLUDE UNKNOWN/NON-FACTUAL CONSTRAINTS IN A SUMMARY(For example, 'There's no specific requirement on X...', 'There's no limit on X...' )
ASK A CLARIFICATION QUESTION BEFORE PROVIDING A SUMMARY TO MAKE SURE YOU HAVE ALL THE CONSTRAINTS AND AN OBJECTIVE FUNCTION.

EXAMPLE OF A SUMMARY:
====
- A coconut seller has to transport coconuts using either rickshaws or ox carts.
- The rickshaws can take 50 coconuts each and cost $10 per trip.
- The ox carts can take 30 coconuts each and cost $8 per trip.
- The seller has at most $200 to spend on transporting the coconuts.
- The number of rickshaws must not exceed the number of ox carts.
====

START THE CONVERSATION WITH A FRIENDLY GREETING, INTRODUCING YOURSELF AND ASKING WHAT THE USER WOULD LIKE TO OPTIMISE. 
\end{lstlisting}
\textbf{QA Agent Prompt:}
\begin{lstlisting}
YOU ARE AGENT IMPERSONATING THE BUSINESS OWNER MENTIONED IN THE PROBLEM STATEMENT(DELIMITED BY ```). 
BE POLITE.
YOU(THE BUSINESS OWNER) ARE TALKING WITH AN EXPERT IN OPTIMIZATIONS.
ACCURATELY PROVIDE INFORMATION AS REQUESTED BASED ON THE PROBLEM STATEMENT. 
MAKE SURE INFORMATION YOU PROVIDING IS CORRECT AND CAN BE FOUND IN THE PROBLEM STATEMENT.
IF THE PROBLEM STATEMENT DOES NOT CONTAIN REQUESTED INFORMATION, SIMPLY SAY YOU DON'T KNOW THESE DETAILS. (for example, "I'm not sure about it, can we skip this")
DO NOT MAKE CALCULATIONS OR INFORMATION MANIPULATING. Use facts from the problem (for example, question: How many X are produced in a day? Answer: I'm not sure, but I know that to produce one X, we need Y minutes.)
DO NOT MENTION THE PROBLEM STATEMENT ANYWHERE; ACT AS IF IT IS YOUR PERSONAL KNOWLEDGE. 

THE PROBLEM STATEMENT:
```
{0}
```

START THE CONVERSATION WITH A WARM GREETING
\end{lstlisting}
\end{document}